\RequirePackage{xcolor}
\documentclass[journal,twoside,web]{ieeecolor}
\usepackage{tuffc}
\usepackage{cite}
\usepackage{pifont}
\usepackage{amsmath,amssymb,amsfonts,multirow,calrsfs,newtxmath}
\usepackage{algorithmic}
\usepackage{graphicx}
\usepackage{subcaption}
\usepackage{multirow}
\usepackage{textcomp}
\usepackage{wrapfig,colortbl}
\usepackage{dsfont}
\usepackage[colorlinks=true, linkcolor=red, citecolor=red, urlcolor=black]{hyperref}
\usepackage{booktabs}
\usepackage{framed}
\usepackage{latexsym}
\usepackage{url}
\usepackage[utf8]{inputenc}

\usepackage{caption,array,booktabs}

\usepackage[titlenumbered,ruled]{algorithm2e}

\newcommand{\cmark}{\ding{51}}
\newcommand{\xmark}{\ding{55}}
\newcolumntype{C}[1]{>{\centering\let\newline\\\arraybackslash\hspace{0pt}}m{#1}}
\usepackage{wrapfig,colortbl}

\definecolor{abstractbg}{rgb}{1,0.969,0.914}
\def\BibTeX{{\rm B\kern-.05em{\sc i\kern-.025em b}\kern-.08em
    T\kern-.1667em\lower.7ex\hbox{E}\kern-.125emX}}
\begin{document}
\title{Grounding DINO-US-SAM: Text-Prompted Multi-Organ Segmentation in Ultrasound with LoRA-Tuned Vision–Language Models}
\author{Hamza Rasaee, Taha Koleilat, Hassan Rivaz, \IEEEmembership{Senior Member, IEEE}
\thanks{H. Rasaee, T. Koleilat, and H. Rivaz are
with the Department of Electrical and Computer Engineering,
Concordia University, Montreal, QC, Canada (e-mail: hamza.rasaee@mail.concordia.ca; taha.koleilat@concordia.ca;hassan.rivaz@concordia.ca).
}
}

\IEEEtitleabstractindextext{%
\fcolorbox{abstractbg}{abstractbg}{%
\begin{minipage}{\textwidth}\rightskip2em\leftskip\rightskip\bigskip
\begin{wrapfigure}[20]{r}{2.5in}%
\hspace{-3pc}\includegraphics[width=2.5in,height=2.5in]
{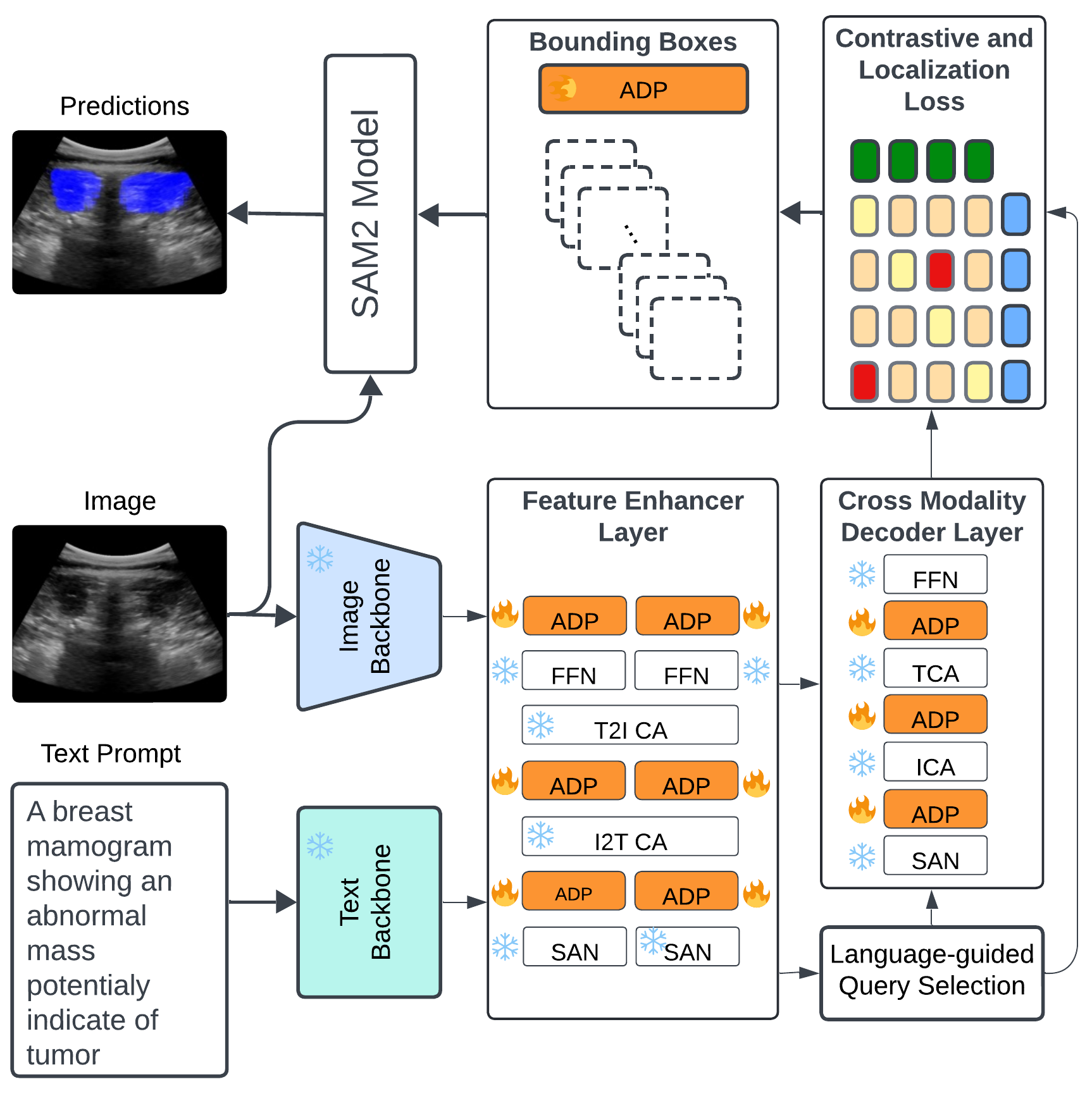}
\end{wrapfigure}
\begin{abstract}
Accurate and generalizable object segmentation in ultrasound imaging remains a significant challenge due to anatomical variability, diverse imaging protocols, and limited annotated data. In this study, we propose a prompt-driven vision-language model (VLM) that integrates Grounding DINO with SAM2 (Segment Anything Model2) to enable object segmentation across multiple ultrasound organs. A total of 18 public ultrasound datasets, encompassing the breast, thyroid, liver, prostate, kidney, and paraspinal muscle, were utilized. These datasets were divided into 15 for fine-tuning and validation of Grounding DINO using Low Rank Adaptation (LoRA) to the ultrasound domain, and 3 were held out entirely for testing to evaluate performance in unseen distributions. Comprehensive experiments demonstrate that our approach outperforms state-of-the-art segmentation methods, including UniverSeg, MedSAM, MedCLIP-SAM, BiomedParse, and SAMUS on most seen datasets while maintaining strong performance on unseen datasets without additional fine-tuning. These results underscore the promise of VLMs in scalable and robust ultrasound image analysis, reducing dependence on large, organ-specific annotated datasets. We will publish our code on \url{code.sonography.ai} after acceptance.
\end{abstract}

\bigskip
\end{minipage}}}

\maketitle

\begin{table*}[!t]
\arrayrulecolor{subsectioncolor}
\setlength{\arrayrulewidth}{1pt}
{\sffamily\bfseries\begin{tabular}{lp{6.75in}}\hline
\rowcolor{abstractbg}\multicolumn{2}{l}{\color{subsectioncolor}{\itshape
Highlights}{\Huge\strut}}\\
\rowcolor{abstractbg}$\bullet$ & A novel segmentation method is proposed by pairing a fine-tuned Grounding DINO with  SAM2 to segment ultrasound images without introducing new complex architectures.\\
\rowcolor{abstractbg}$\bullet${\large\strut} & Enables free-form natural-language prompts without any clicks, boxes, or organ-specific retraining.\\
\rowcolor{abstractbg}$\bullet${\large\strut} &
The method outperforms UniverSeg, MedSAM, MedCLIP-SAM, BiomedParse, and SAMUS on breast, liver, and prostate datasets.\\
\rowcolor{abstractbg}$\bullet${\large\strut} & Demonstrates strong generalization to unseen datasets with varying similarity levels: breast (BUSBRA, similar), thyroid (TNSCUI, moderately similar), and paraspinal muscles (LUMINOUS, dissimilar) domains.\\
\rowcolor{abstractbg}$\bullet${\large\strut} & Utilizes parameter-efficient fine-tuning via Low-Rank Adaptation (LoRA), allowing effective model adaptation with minimal computational overhead.\\[2em]\hline
\end{tabular}
}
\setlength{\arrayrulewidth}{0.4pt}
\arrayrulecolor{black}
\end{table*}

\section{Introduction}
\label{sec:introduction}
Ultrasound imaging is extensively used in clinical practice due to its safety, affordability, portability, and real-time capabilities. It plays a vital role in cancer screening, disease staging, and image-guided interventions across various anatomies, including the breast, thyroid, liver, prostate, kidney, and musculoskeletal system. Despite these advantages, ultrasound imaging presents intrinsic challenges that complicate automated analysis. Issues like low tissue contrast, speckle noise, acoustic shadowing, and operator-dependent variability degrade image quality and hinder the precise delineation of anatomical structures, ultimately affecting automated segmentation algorithms' performance and generalizability.

To overcome these limitations, deep learning (DL) methods have become standard in ultrasound image analysis. In fully supervised DL approaches, the models completely learn from the labeled data. Architectures such as convolutional neural networks (CNNs) effectively capture local features~\cite{amiri2020fine}, while vision transformer-based (ViT) models capture long-range dependencies and global context via self-attention. Additionally, hybrid designs combining CNNs and transformers have improved performance across many ultrasound segmentation tasks. For example, He  \textit{et al.} introduced HCTNet~\cite{he2023hctnet}, a hybrid CNN-transformer-based layers for breast lesion segmentation, while Xu \textit{et al.} proposed MCV-UNet~\cite{xu2024mcv} that used a CNN-ViT backbone for multiscale nerve segmentation. For fetal head estimation, Jiang \textit{et al.}~\cite{jiang2024intrapartum} proposed a dual student-teacher model. In musculoskeletal imaging, Chen \textit{et al.}~\cite{chen2024hybrid} designed a hybrid transformer tailored for arm segmentation, effectively capturing both global and local anatomical features. Models such as MicroSegNet~\cite{jiang2024microsegnet}, LightBTSeg~\cite{guo2023lightbtseg}, and AAU-Net~\cite{chen2022aau} have achieved state-of-the-art (SOTA) performance on breast and prostate ultrasound datasets. Additionally, UniverSeg~\cite{butoi2023universeg} trains across multiple medical imaging modalities using a shared encoder and conditioning mechanism, supporting cross-task generalization without structural changes. However, it requires a training set size of 16 segmented sample images to adapt to new tasks.

Despite the improvements that these fully supervised models brought to the field, they rely on high-quality annotations. To reduce annotation demands, semi-supervised learning (SSL) methods have recently been explored~\cite{chen2024striving}, including consistency-based SSL for muscle ultrasound~\cite{rasaee2025ultrasound}. Unsupervised techniques such as deep spectral learning~\cite{tmenova2024deep} further minimize manual labeling requirements. While SSL-based and unsupervised models reduce the reliance on annotated ultrasound data, shape-guided architectures like SHAN~\cite{zhang2024shan} tackle distribution shifts in multi-center thyroid ultrasound segmentation and effectively leverage large-scale annotated datasets from diverse imaging centers. These models are organ-specific and require extensive retraining or architectural modifications to generalize across different organs, with additional retraining needed for each new target organ.

Despite the numerous advancements mentioned above, such as reducing annotation needs or enhancing robustness across different anatomical structures, these methods often address isolated challenges and fall short of fulfilling broader clinical and technical requirements. A universally generalizable solution that performs reliably across a wide range of tasks, organs, and imaging conditions remains elusive. This has fueled growing interest in more flexible and adaptable paradigms, such as prompt-driven segmentation using vision-language models (VLMs). VLMs offer several advantages: they enable zero-shot generalization to previously unseen anatomical structures through text prompts~\cite{radford2021learning, koleilat2024medclip}, allow seamless adaptation across different segmentation tasks without task-specific retraining~\cite{liu2024grounding}, and promote scalability across imaging modalities and clinical use cases~\cite{wang2022medclip}.

Prompt-driven segmentation using VLM approaches has emerged as a promising alternative for generalizing to previously unseen anatomical structures, enabling segmentation without the need for additional training~\cite{ravi2024sam}. Building on this paradigm, several models have been proposed to leverage the strengths of VLM integration for more generalizable medical image segmentation \ref{tab:litrature}. For example, Grounding DINO~\cite{liu2024grounding} is one of the widely used VLM approaches that performs open-set object detection guided by input prompts. By combining image features with text embeddings, it enables precise localization of objects described in free-form text, making it one of the leading approaches for language-guided visual understanding. BiomedParse~\cite{zhao2025foundation} demonstrates the feasibility of joint segmentation, detection, and recognition across multiple modalities, reinforcing the potential of foundation models. Variants such as MedCLIP-SAM~\cite{koleilat2024medclip} and MedCLIP-SAM2~\cite{koleilat2024medclipsamv2} integrate CLIP-style~\cite{radford2021learning} image-text pre-training with segmenta anything model (SAM)~\cite{ma2024segment} for CT and MRI segmentation. However, their effectiveness on ultrasound data remains limited due to modality mismatches.

In the ultrasound domain, Ferreira \textit{et al.}~\cite{ferreira2025segment} proposed a prompt-based method for small structure segmentation, leveraging iterative point prompts and image transformations. UltraSAM~\cite{meyer2024ultrasam} enhances SAM’s sensitivity to ultrasound texture yet lacks vision-language interaction and remains task-specific. SAM-MedUS~\cite{tian2025sam} demonstrates strong generalization across ultrasound tasks through multi-domain training; however, its reliance on manually provided prompts introduces limitations to automation and scalability in clinical workflows. Interactive models like ClickSAM~\cite{guo2024clicksam} address the automation limitations of manual prompting by reducing annotation effort through click-based inputs, but they still rely on spatial interactions, which impose another layer of constraint. 

SAMUS~\cite{lin2024beyond} incorporates a modified ViT encoder, parallel CNN branch, and cross-branch attention to improve segmentation efficiency. Although SAMUS provides a promising approach for automatic ultrasound segmentation, recent experiments have shown that it relies on accurate user-defined point prompts~\cite{yin2025apg}, which limits its practicality for novice users who lack expertise with ultrasound imaging and human anatomy.  Gowda \textit{et al.}~\cite{gowda2024cc} utilized ChatGPT to generate contextual prompts for SAMUS, enhancing its ability to capture subtle ultrasound features; however, it was not fine-tuned for the ultrasound domain to avoid the computationally expensive fine-tuning task.

SAM requires user-provided spatial prompts, such as bounding boxes, which limits full automation. To address this, APG-SAM~\cite{yin2025apg} recently introduced automatic spatial prompting through detection networks, eliminating the need for manual bounding box inputs. Although this is a promising advancement, the method lacks explicit semantic guidance through text and is limited to breast ultrasound images, reducing its generalizability to other anatomies. Taking a different approach, Chen \textit{et al.}~\cite{chen2025multi} developed MOFO, a multi-organ ultrasound segmentation model guided by anatomical priors. Although an important step forward, it only supports fixed organ-specific task prompts. In other words, the prompts are hard-coded to a closed list of organs and are not flexible~\cite{chen2025multi}. 
In addition, this work relies on learning from multiple organs simultaneously, requiring the presence of data from various organs in the training set for effective segmentation. It also doesn't provide inference time, potentially limiting real-time clinical deployment.  

To address the limitations of manual prompting in SAM-based segmentation, we incorporate Grounding DINO~\cite{liu2024grounding} as a prior model to automatically generate the bounding boxes needed for SAM-based segmentation of ultrasound images. Grounding DINO integrates cross-modal attention between image and text features to generate bounding boxes from input prompts without requiring predefined classes or user interaction. This capability enables automatic spatial prompting for SAM, allowing it to segment anatomical structures in ultrasound images based on high-level semantic cues such as organ names. Unlike prior methods like APG-SAM, which rely solely on visual detection networks and are restricted to specific anatomies (e.g., breast), our approach combines the semantic flexibility of text-based localization with the generalization ability of VLMs, resulting in a more scalable and fully automated segmentation pipeline across diverse ultrasound imaging domains. To this end, our proposed approach integrates a Low-Rank Adaptation (LoRA), tuned Grounding DINO detector, with a frozen SAM2 decoder. Notably, only 1.7\% of the Grounding DINO parameters are updated during training, while the weights of SAM2 remain entirely frozen. Therefore, the proposed LoRA-based grounding strategy overcomes the geometric-only guidance of APG-SAM\cite{yin2025apg} and the rigid organ-specific prompt design of Chen \textit{et al.}\cite{chen2025multi}, offering a scalable, annotation-free solution for ultrasound segmentation in diverse clinical applications. 
For natural images, Das~\textit{et al.}~\cite{das2024prompting} demonstrated the effectiveness of prompting foundational models for omni-supervised instance segmentation, showcasing how models like Grounding DINO and SAM can be adapted to complex instance segmentation tasks.
Chen \textit{et al.}~\cite{chen2025interpreting} presented a visual precision search interpretability framework built on top of Grounding DINO and SAM to study object-level model behavior. Zhang~\textit{et al.}~\cite{zhang2024efficientvit} introduced EfficientViT‑SAM, which utilizes Grounding DINO to provide bounding box proposals to SAM in order to generate segmentation masks in a zero-shot open-world setting.

The contribution of the proposed approaches can be summarized as follows:

\begin{itemize}
\item Accepts free-form natural-language prompts (e.g. \texttt{"tumor", "malignant lesion"}) without any clicks, boxes, or organ-specific retraining

\item Achieves SOTA DSC score on 18 public ultrasound datasets covering 6 organ systems

\item Generalises to unseen anatomies and acquisition protocols

\item Runs in real time (0.33 s  for 800 × 800 images on an affordable Titan V GPU released in 2017), facilitating clinical deployment.
\end{itemize}

\begin{table*}[ht]
    \centering
    \caption{Comparison of recent prompt-driven segmentation methods on ultrasound imaging.  Point and bounding box columns indicate whether the method needs \emph{manual spatial prompts}. Our proposed method supports fully automated segmentation with text-based interactions, overcoming common modality mismatches and manual prompting requirements.}

    \label{tab:litrature}
    \renewcommand{\arraystretch}{1.2}
    \begin{tabular}{lccccccc}
        \toprule
        \textbf{Model} & \textbf{Ultrasound-based} & \textbf{Text prompt} & \textbf{Point} & \textbf{Bounding box} & \textbf{Notes}\\
        \midrule
        Grounding DINO & \xmark & \cmark & \xmark & \xmark & Developed for natural images\\
        BiomedParse & \xmark & \cmark & \xmark & \xmark & Not optimized for US\\
        MedCLIP-SAM & \xmark & \cmark & \xmark & \xmark & Limited US performance\\
        UltraSAM & \cmark & \xmark & \cmark & \cmark & No text interaction\\
        SAM-MedUS & \cmark & \xmark & \cmark & \cmark & Requires manual spatial prompts\\
        ClickSAM & \xmark & \xmark & \cmark & \xmark & Spatial interaction needed\\
        SAMUS & \cmark & \xmark & \cmark & \xmark & Needs expert points\\
        APG-SAM & \cmark & \xmark & \xmark & \xmark & Limited to breast US\\
        MOFO & \cmark & \xmark & \xmark & \xmark & Closed organ list\\
        \textbf{Proposed} & \textbf{\cmark} & \textbf{\cmark} & \textbf{\xmark} & \textbf{\xmark} & Optimized for US; no manual spatial prompts needed \\
        \bottomrule
    \end{tabular}
\end{table*}

\section{Method}
\subsection{Dataset Preparation}

A total of 18 public ultrasound datasets covering various organs, including breast, thyroid, liver, prostate, kidney, and muscle, were used in this study, as summarized in Table~\ref{tab:datasets}. These datasets provide diverse anatomical structures, acquisition protocols, and imaging conditions, enhancing the robustness and generalizability of the proposed method.

For training and validation, 15 datasets (referred to as the seen set) were used, consisting of 12,924 images for training and 3,881 for validation. Key datasets include BrEaST (breast)~\cite{pawlowska2024curated}, BUID (breast)~\cite{ardakani2023open, hamyoon2022artificial, homayoun2022applications}, BUSUC (breast)~\cite{bus_uc}, BUSUCML (breast)~\cite{vallez2025bus}, BUSB (breast)~\cite{yap2017automated}, BUSI (breast)~\cite{al2020dataset}, STU (breast)~\cite{STUHospital2024}, S1 (thyroid)~\cite{guo2021segmentation}, TN3K (thyroid)~\cite{gong2021multi}, TG3K (liver)~\cite{gong2022less}, 105US (liver)~\cite{hann2017algorithm}, AUL (prostate)~\cite{xu2023improving}, MicroSeg (prostate)~\cite{le2025u2}, RegPro (breast)~\cite{hamid2019smarttarget}, and kidneyUS (kidney)~\cite{singla2023open}.

To assess generalization under domain shifts, three additional datasets, BUSBRA (breast)~\cite{gomez2024bus}, TNSCUI (thyroid)~\cite{zhou2020thyroid}, and Luminous (back muscle)~\cite{belasso2020luminous}, were used exclusively for testing. These datasets (highlighted in gray in Table~\ref{tab:datasets}) comprise 2,808 images and are considered out-of-distribution (i.e., never seen during training or validation), enabling evaluation on previously unseen domains.

For all datasets, ground truth segmentation masks were converted into tight bounding boxes by extracting the minimum enclosing rectangles around annotated regions. Each image was paired with a textual prompt derived (e.g., ``benign'' or ``malignant'') during training. All images were resized to $800 \times 800$ pixels, and their corresponding bounding boxes were rescaled accordingly. The unseen datasets were reserved exclusively for inference.

\begin{table}[ht]
    \centering
    \caption{Public ultrasound datasets used in this study and their distribution across train, validation, and test sets. Three datasets used exclusively for testing (not seen during training or validation) are highlighted in gray.  The only exception is the UniverSeg baseline, which requires a 16-image support set. Therefore, 16 manually segmented images from each unseen dataset were provided for this method.}
    \label{tab:datasets}
    \renewcommand{\arraystretch}{1.2}
    \begin{tabular}{clccccc}
    \toprule
    \textbf{Organ} & \textbf{Dataset} & \textbf{Total} & \textbf{Train} & \textbf{Val} & \textbf{Test} \\
    \midrule
    \multirow{9}{*}{Breast} 
        & BrEaST \cite{pawlowska2024curated}               & 252  & 176 & 50  & 26 \\
        & BUID \cite{ardakani2023open, hamyoon2022artificial, homayoun2022applications} & 233  & 161 & 46  & 26 \\
        & BUSUC \cite{bus_uc}                               & 810  & 566 & 161 & 83 \\
        & BUSUCML \cite{vallez2025bus}                      & 264  & 183 & 52  & 29 \\
        & BUSB \cite{yap2017automated}                      & 163  & 114 & 32  & 17 \\
        & BUSI \cite{al2020dataset}                         & 657  & 456 & 132 & 69 \\
        & STU \cite{STUHospital2024}                                              & 42   & 29  & 8   & 5 \\
        & S1 \cite{guo2021segmentation}                     & 202  & 140 & 40  & 22 \\
        \rowcolor{gray!20}
        & BUSBRA \cite{gomez2024bus}                       & 1875 & \textbf{---}  & \textbf{---}   & 1875 \\
    \midrule
    \multirow{3}{*}{Thyroid} 
        & TN3K \cite{gong2021multi}                         & 3493 & 2442 & 703 & 348 \\
        & TG3K \cite{gong2022less}                          & 3565 & 2497 & 713 & 355 \\
        \rowcolor{gray!20}
        & TNSCUI \cite{zhou2020thyroid}                    & 637  & \textbf{---}    & \textbf{---} & 637 \\
    \midrule
    \multirow{3}{*}{Liver} 
        & 105US \cite{hann2017algorithm}                    & 105  & 73   & 21  & 11 \\
        & AUL \cite{xu2023improving}                         & 533  & 351  & 120 & 62 \\
    \midrule
    \multirow{2}{*}{Prostate} 
        & MicroSeg \cite{le2025u2}                           & 2283 & 1527 & 495 & 261 \\
        & RegPro \cite{hamid2019smarttarget}                 & 4218 & 2952 & 843 & 423 \\
    \midrule
    Kidney & kidneyUS \cite{singla2023open}                      & 1963 & 1257 & 465 & 241 \\
    \midrule
    \rowcolor{gray!20}
        Back Muscle
        & Luminous \cite{belasso2020luminous}               & 296  & \textbf{---}   & \textbf{---}   & 296 \\
    \midrule
     & \textbf{Total} & \textbf{18783} & \textbf{12924} & \textbf{3881} & \textbf{1978} \\
    \bottomrule
    \end{tabular}
\end{table}

\subsection{Model Architecture and LoRA-Based Fine-Tuning}

\begin{figure*}[ht]
    \centering
    \includegraphics[width=\textwidth]{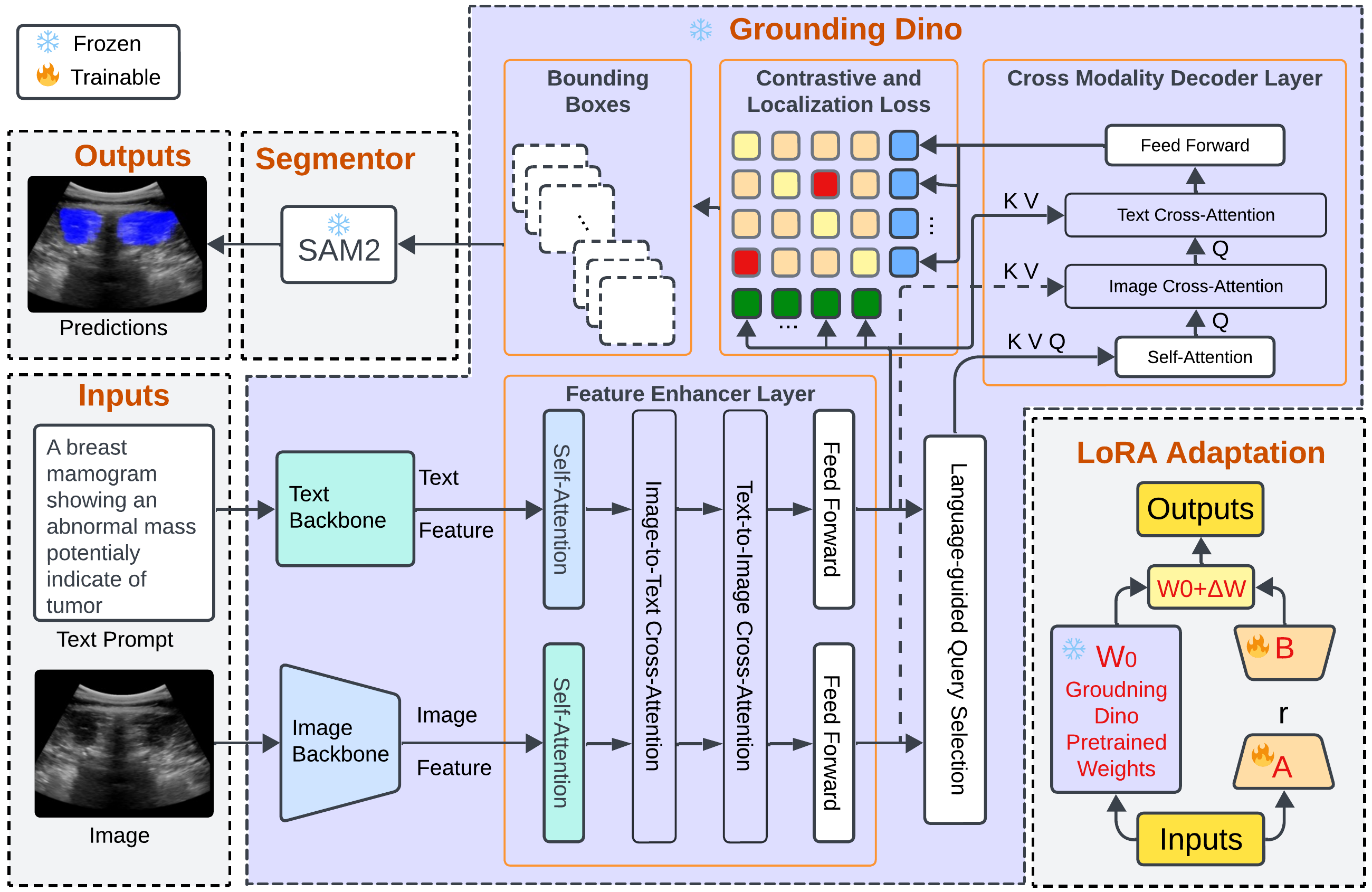} 
    \caption{The system uses frozen Grounding DINO and SAM2 for zero-shot ultrasound segmentation via natural language prompts. Image and text features are extracted, fused with attention layers, and used to predict bounding boxes. SAM2 performs segmentation based on these localized boxes.}
    
    \label{fig:grounding_lora}
\end{figure*}

As illustrated in Fig.~\ref{fig:grounding_lora}, the proposed framework builds on Grounding DINO, a transformer-based vision-language model (VLM) that performs joint image-text grounding. 

LoRA introduces low-rank trainable matrices to approximate weight updates while keeping pretrained weights frozen. Given a weight matrix $W_0 \in \mathbb{R}^{d \times k}$, LoRA represents the update as:
\[
W = W_0 + \Delta W = W_0 + BA,
\]
where $A \in \mathbb{R}^{r \times k}$ and $B \in \mathbb{R}^{d \times r}$ are learnable matrices with rank $r \ll \min(d, k)$, significantly reducing the number of trainable parameters. \( B \) is initialized to zero and \( A \) is randomly initialized, so that \( BA = 0 \) at the start of training. This ensures that the initial output of the model matches the pretrained model.

It consists of a frozen image backbone, a frozen BERT-based text encoder, and a transformer-based encoder-decoder structure~\cite{liu2024grounding}. To efficiently adapt this model to ultrasound data while preserving pretrained knowledge, we introduce Low-Rank Adaptation (LoRA)~\cite{hu2022lora} into selected modules:

\textbf{Encoder (Feature Enhancer):} LoRA adapters are inserted into the feed-forward and cross-attention layers within the feature enhancer, enabling task-specific adaptation for both image and text features.

\textbf{Decoder (Cross-Modality):} LoRA modules are injected into all decoder layers, specifically targeting sampling offsets, attention weights, projection layers, and cross-modal attention, enhancing spatial grounding aligned with the textual prompt.

\textbf{Text Encoder:} While the BERT-based text encoder remains frozen, LoRA modules are selectively applied to the self-attention output projections and feed-forward layers, introducing controlled flexibility for medical domain text adaptation.

\textbf{Other Components:} The feature map layer bridging visual and textual features is adapted via LoRA, while the bounding box regression head is left fully trainable without LoRA, allowing effective localization without disrupting pretrained representations.

Bounding box predictions, aligned through contrastive and localization losses, guide the downstream \textbf{SAM2} model to generate fine-grained segmentation masks from ultrasound inputs. This setup makes only $\approx$1.7\% of the total trainable parameters of Grounding DINO, enabling efficient fine-tuning with low GPU memory usage and reduced overfitting risk, which is particularly important in low-data ultrasound imaging scenarios.

\subsection{Loss Function}
The proposed model is trained using a composite loss function that combines three components to jointly supervise bounding box regression, spatial alignment, and text-guided classification:

\begin{equation}
\mathcal{L}_{\text{total}} = \lambda_{\text{L1}} \mathcal{L}_{\text{L1}} + \lambda_{\text{GIoU}} \mathcal{L}_{\text{GIoU}} + \lambda_{\text{focal}} \mathcal{L}_{\text{focal}}
\end{equation}

The loss components are defined as follows:

\vspace{1mm}
\textbf{1. $\mathcal{L}_{\text{L1}}$: Bounding Box Regression} \\
This is the standard L1 loss between predicted box coordinates $\hat{b} = (x, y, w, h)$ and ground truth $b$:

\begin{equation}
\mathcal{L}_{\text{L1}} = \lVert \hat{b} - b \rVert_1
\end{equation}

\vspace{1mm}
\textbf{2. $\mathcal{L}_{\text{GIoU}}$: Generalized IoU Loss} \\
To evaluate the alignment between predicted and ground truth bounding boxes, we adopt the Generalized Intersection over Union (GIoU) metric~\cite{rezatofighi2019generalized}, which extends the standard IoU by incorporating shape and distance information. Given a predicted box $B_p$ and a ground truth box $B_{gt}$, the GIoU is defined as:

\begin{equation}
\mathrm{GIoU}(B_p, B_{gt}) = \mathrm{IoU}(B_p, B_{gt}) - \frac{|C \setminus (B_p \cup B_{gt})|}{|C|}
\end{equation}
where
\begin{equation}
\mathrm{IoU}(B_p, B_{gt}) = \frac{|B_p \cap B_{gt}|}{|B_p \cup B_{gt}|}
\end{equation}
is the standard Intersection over Union, and $C$ is the smallest enclosing box that contains both $B_p$ and $B_{gt}$. This formulation penalizes predictions that are far from the ground truth, even if they have partial overlap, leading to better convergence and improved spatial accuracy during training.

\vspace{1mm}
\textbf{3. $\mathcal{L}_{\text{focal}}$: Focal Contrastive Classification Loss} \\
To enable open-vocabulary grounding, we adopt a contrastive version of the Focal Loss~\cite{lin2017focal}. Let $q_i$ be the $i$-th decoder query and $t_j$ the $j$-th token embedding in the text prompt. The model computes similarity scores via scaled dot products:

\begin{equation}
\text{logits}_{i,j} = \frac{q_i^\top t_j}{\tau}
\end{equation}

where $\tau$ is a learned temperature parameter. For each query, we apply focal loss over the logits:

\begin{equation}
\mathcal{L}_{\text{focal}} = -\alpha (1 - p_t)^\gamma \log(p_t)
\end{equation}

where $p_t$ is the softmax probability assigned to the correct token, $\alpha = 0.25$ is the class-balancing factor, and $\gamma = 2.0$ is the focusing parameter (as in the original formulation).

\vspace{1mm}
The final loss is a weighted sum of the above components, with the following coefficients used in all experiments:

\[
\lambda_{\text{L1}} = 0.5, \quad \lambda_{\text{GIoU}} = 0.1, \quad \lambda_{\text{focal}} = 0.1
\]

This formulation enables the model to localize and classify objects based on free-form text prompts with high precision, while preserving generalization to unseen anatomical structures and phrasing.

\subsection{Evaluation Metrics}

Segmentation performance was evaluated using two standard overlap-based metrics: the Intersection over Union (IoU) and the Dice Score Coefficient (DSC). These metrics were computed by comparing the SAM2-generated segmentation masks with the ground truth annotations. IoU measures the overlap between the predicted mask $P$ and the ground truth mask $G$:

\begin{equation}
\text{IoU} = \frac{|P \cap G|}{|P \cup G|}
\end{equation}

\noindent where $|P \cap G|$ is the number of pixels common to both predicted and ground truth masks, and $|P \cup G|$ is the total number of pixels in either mask. IoU provides a strict evaluation of region overlap and penalizes false positives and false negatives equally~\cite{everingham2010pascal}.

DSC, also known as the F1-score in segmentation tasks, measures the harmonic mean of precision and recall:

\begin{equation}
\text{DSC} = \frac{2|P \cap G|}{|P| + |G|}
\end{equation}

This metric emphasizes pixel-wise agreement between $P$ and $G$ and is especially sensitive to small object segmentation~\cite{dice1945measures}. Together, IoU and DSC provide complementary insights into segmentation performance, balancing region overlap and boundary agreement.

\subsection{Implementation Details}

All training was performed on an NVIDIA RTX 4090 GPU (24 GB RAM), and inference speed was benchmarked on a more affordable NVIDIA Titan V released in 2017 (12 GB RAM). The model was trained using 15 public ultrasound datasets across five organs (breast, thyroid, liver, prostate, and kidney). Each dataset was split into training (70\%), validation (20\%), and test (10\%), with the test sets withheld during model development to ensure evaluation on unseen data. 

LoRA fine-tuning was conducted using a batch size of 4 and a learning rate of $5 \times 10^{-4}$ with the AdamW optimizer~\cite{loshchilov2017decoupled} with weight decay of $1 \times 10^{-4}$. Only LoRA parameters were updated, whereas all other weights remained frozen. The training loss included focal loss for classification, L1 and GIoU for localization, and a contrastive alignment loss to optimize image-text correspondence. Early stopping was employed with a patience of 20 epochs based on validation loss to mitigate overfitting. 
All datasets were pooled during training, and data augmentations such as random flipping (with a probability of 50\%), resizing (scale range: 0.8x to 1.2x), padding (up to 10\% of the image size), erasing (with a probability of 20\%, area ratio: 2–10\%), and cropping (covering 80–100\% of the original image) were applied. This led to a single, general-purpose model that was evaluated on all datasets (seen and unseen).

\section{Results}
This section presents quantitative and qualitative evaluations of the proposed method on various ultrasound segmentation tasks. We include ablation studies, comparisons with SOTA methods on seen datasets, and performance analysis on unseen domains to validate the approach's generalizability.

\subsection{Ablation Study}

Table~\ref{tab:ablation} summarizes the ablation study evaluating different configurations of the Grounding DINO model combined with various segmentation heads. The baseline configuration, Grounding DINO with SAM2 (trained on natural images), shows limited segmentation performance across all datasets. Fine-tuning Grounding DINO paired with MedSAM notably improves both DSC and IoU scores. However, the best performance is achieved by fine-tuning Grounding DINO paired with SAM2, which consistently outperforms the other configurations across breast, liver, and prostate datasets.
The baseline configuration (without fine-tuning) performs the worst, especially on complex datasets like AUL, where it achieves only $13.77\%$ DSC. Fine-tuning with MedSAM improves segmentation quality across all organs, but the most consistent and substantial improvements are observed when fine-tuning with SAM2. This configuration yields the best DSC and IoU scores in all datasets. For instance, on the BrEaST dataset, the DSC improves from $23.87\%$ (baseline) to $76.65\%$ when using the fine-tuned SAM2. Similarly, the MicroSeg prostate dataset shows a leap from $61.44\%$ to $88.56\%$ in DSC, with a corresponding increase in IoU to $81.17\%$. Even in the challenging AUL dataset, performance climbs to $44.02\%$ DSC, a significant gain over other methods. Statistical significance was assessed via paired \textit{t}-tests across test samples. We mark results with $p < 0.005$ and $p < 0.01$ as \textbf{**} and \textbf{*}, respectively. These tests confirm that the observed gains are not due to chance, particularly for breast datasets, where the differences are consistently significant.
Given its consistent superiority in both mean performance and statistical robustness, we adopt fine-tuned Grounding DINO + SAM2 for all subsequent experiments.

\begin{table*}[ht]
    \centering
    \caption{Ablation study comparing Grounding DINO's segmentation performance (DSC and IoU scores) paired with different segmentation heads across multiple ultrasound datasets. The evaluated configurations include: (1) Grounding DINO + SAM2 (trained on natural images), (2) Fine-tuned Grounding DINO + MedSAM, and (3) Fine-tuned Grounding DINO + SAM2. Results are reported as mean $\pm$ standard deviation. Bolded results indicate the highest score in each comparison. The third method consistently outperforms other methods across all datasets and organs, and is used in the rest of the experiments. Statistical significance was assessed using paired t-tests: $^{**}$  denotes $p < 0.005$ and $^{*}$ denotes $p < 0.01$.}
    
    \label{tab:ablation}
    \renewcommand{\arraystretch}{1.2}
    \begin{tabular}{llcccccc}
        \toprule
        \multirow{2}{*}{\textbf{Organ}} & \multirow{2}{*}{\textbf{Dataset}} & 
        \multicolumn{2}{c}{\textbf{Grounding DINO + SAM2}} & 
        \multicolumn{2}{c}{\textbf{Fine-tuned Grounding DINO  + MedSAM}} & 
        \multicolumn{2}{c}{\textbf{Fine-tuned Grounding DINO + SAM2}} \\
        \cmidrule(lr){3-4} \cmidrule(lr){5-6} \cmidrule(lr){7-8}
        & & \textbf{DSC (\%) $\uparrow$} & \textbf{IoU (\%) $\uparrow$} & \textbf{DSC (\%) $\uparrow$} & \textbf{IoU (\%) $\uparrow$} & \textbf{DSC (\%) $\uparrow$} & \textbf{IoU (\%) $\uparrow$} \\

            \midrule
            \multirow{3}{*}{Breast} 
            & BrEaST & $23.87±30^{**}$ & $18.08±27^{**}$ & $49.62±28^{**}$ & $37.33±24^{**}$ & \textbf{76.65±23} & \textbf{66.24±23}\\
            & BUID & $44.09±35^{**}$ & $36.08±35^{**}$ &  $61.43±29^{*}$ & $49.38±25^{**}$ & \textbf{82.92±24} & \textbf{75.83±25}\\
            & BUSI & $29.55±34^{**}$ & $23.31±31^{**}$ &  $51.35±30^{**}$ & $39.74±26^{**}$ & \textbf{77.79±26} & \textbf{69.29±27}\\
            \midrule
            \multirow{1}{*}{Liver} 
            & AUL &  $13.77±13^{**}$ & $7.94±8^{**}$ &  $32.50±28$ & $23.25±23^{**}$ & \textbf{44.02±36} & \textbf{35.79±33}\\
            \midrule
            \multirow{1}{*}{Prostate} 
            & MicroSeg & $61.44±21^{**}$ & $47.47±21^{**}$ &  $56.84±22^{**}$ & $42.75±20^{**}$ & \textbf{88.56±13} & \textbf{81.17±15}\\

        \bottomrule
    \end{tabular}
\end{table*}

\subsection{Comparison on Seen Datasets}

A comprehensive comparison with recent SOTA segmentation methods on seen datasets is provided in Table~\ref{tab:seen}. The proposed method outperforms MedCLIP-SAM, MedCLIP-SAMv2, UniverSeg, SAMUS, and BiomedParse on most datasets. For example, our method on the BUSUC breast dataset achieves the highest DSC score of $91.68\%$ and IoU of $85.16\%$, surpassing all baselines. Similar improvements are observed on BUSB, S1, TN3K, TG3K, 105US, and AUL datasets, confirming the robustness and effectiveness of the proposed approach on seen domains.

Notably, some of the compared methods, such as BiomedParse and SAMUS, benefited from significantly larger pre-training datasets (BiomedParse used approximately 3.4 million medical images, and SAMUS utilized 30,000 ultrasound images). Despite this, our method, trained on a substantially smaller dataset, consistently achieves superior performance. It is also important to note that because the model for AutoSAMUS was not released, random points within the ground truth mask were provided during inference to SAMUS. In other words, the results of this method are not fully automatic and are obtained by providing the model with a point prompt. In contrast, our method operates with no spatial prompts, making the comparison conservative and highlighting its robustness.

\setlength{\tabcolsep}{2pt}
\begin{table*}[ht]
    \centering
    \scriptsize
    \caption{Quantitative comparison of segmentation performance (DSC and IoU scores) on publicly available ultrasound datasets among SOTA methods: UniverSeg, BiomedParse, SAMUS, MedCLIP-SAM, MedCLIP-SAMv2, and the proposed method. Results are reported as mean $\pm$ standard deviation. Bolded results indicate the highest score in each comparison.
    Statistical significance was assessed using paired t-tests: $^{**}$  denotes $p < 0.005$ and $^{*}$ denotes $p < 0.01$.}
    
    \label{tab:seen}
    \renewcommand{\arraystretch}{1.2}
    \begin{tabular}{llcccccccccccc}
        \toprule
        \multirow{2}{*}{\textbf{Organ}} & \multirow{2}{*}{\textbf{Dataset}} & 
        \multicolumn{2}{c}{\textbf{UniverSeg\cite{butoi2023universeg}}} &
        \multicolumn{2}{c}{\textbf{BiomedParse\cite{zhao2025foundation}}} &
        \multicolumn{2}{c}{\textbf{SAMUS}} &
        \multicolumn{2}{c}{\textbf{MedCLIP-SAM}} & 
        \multicolumn{2}{c}{\textbf{MedCLIP-SAMv2}} & 
        \multicolumn{2}{c}{\textbf{Ours}} \\
        \cmidrule(lr){3-4} \cmidrule(lr){5-6} \cmidrule(lr){7-8} \cmidrule(lr){9-10} \cmidrule(lr){11-12}\cmidrule(lr){13-14}
        & & \textbf{DSC (\%) $\uparrow$} & \textbf{IoU (\%) $\uparrow$} & \textbf{DSC (\%) $\uparrow$} & \textbf{IoU (\%) $\uparrow$} & \textbf{DSC (\%) $\uparrow$} & \textbf{IoU (\%) $\uparrow$} & \textbf{DSC (\%) $\uparrow$} & \textbf{IoU (\%) $\uparrow$} & \textbf{DSC (\%) $\uparrow$} & \textbf{IoU (\%) $\uparrow$} & \textbf{DSC (\%) $\uparrow$} & \textbf{IoU (\%) $\uparrow$}  \\
        
\midrule
\multirow{8}{*}{Breast} 
& BrEaST &  $60.30±23$ & $47.01±24^{**}$ &  $76.36±20$ & $65.19±22$ & $60.78±22$ & $46.95±21$ & $65.02±21$ & $51.18±19^{**}$ & $72.91±25$ & $62.24±25$ & \textbf{76.65±23} & \textbf{66.24±23} \\
& BUID & $56.30±30^{**}$ & $45.27±29^{**}$ &  $82.77±21$ & $74.54±23$ & $71.74±17$ & $57.93±16$ & $70.13±22^{*}$ & $57.90±24^{**}$ & $78.48±23$ & $69.48±26$ & \textbf{82.92±24} & \textbf{75.83±25} \\
& BUSUC & $51.94±28^{**}$ & $39.89±26^{**}$ &  $87.17±12^{**}$ & $78.92±16^{**}$ & $71.46±17^{**}$ & $58.06±19^{**}$ &  $80.66±11^{**}$ & $68.88±14^{**}$ &  $87.40±11^{**}$ & $78.99±15^{**}$ &  \textbf{91.68±6} & \textbf{85.16±9} \\
& BUSUCLM & $52.96±22^{**}$ & $39.19±21^{**}$ & $66.34±31$ & $56.50±30$ & $44.34±32^{**}$ & $34.32±29^{**}$ & $56.88±21^{**}$ & $42.48±19^{**}$ & $66.09±27$ & $54.66±27$ & \textbf{76.74±24} & \textbf{66.75±24} \\
& BUSB & $82.58±15$ & $72.87±20$ & $73.93±31$ & $66.12±31$ & $64.40±20^{**}$ & $50.52±20^{**}$ & $73.84±23^{**}$ & $62.84±24^{**}$ & $82.99±23 $& $75.46±23$ & \textbf{91.04±8} & \textbf{84.30±11} \\
& BUSI & $43.19±26^{**}$ & $31.16±22^{**}$ & \textbf{89.39±14\textsuperscript{**}} & \textbf{82.72±15\textsuperscript{**}} & $82.45±10^{**}$ & $71.19±13^{**}$ & $56.64±30^{**}$ & $45.51±29^{**}$ & $63.60±34^{**}$ & $54.40±32^{**}$ & $77.79±26$ & $69.29±27$ \\
& STU & $64.73±17$ & $50.16±19$ & $87.07±5$ & $77.50±8^{**}$ &  $68.46±11$ & $53.10±12$ & $83.77±9$ & $73.03±13$ & \textbf{90.11±6\textsuperscript{**}} & \textbf{82.56±10\textsuperscript{**}} & $87.89±6$ & $78.87±9$ \\
& S1 & $57.30±25^{**}$ & $44.25±24^{**}$ &  $93.28±4$ & $87.63±6$ & $76.09±20$ & $64.54±20$ & $78.92±9^{**}$ & $66.02±12^{**}$ & $88.04±8^{**}$ & $79.40±11^{**}$ & \textbf{93.37±5\textsuperscript{**}} & \textbf{87.96±8\textsuperscript{**}} \\
\midrule
\multirow{2}{*}{Thyroid} 
& TN3K & $39.46±21^{**}$ & $26.88±18^{**}$ & $56.96±34^{**}$ & $47.07±31^{**}$ & \textbf{80.96±12\textsuperscript{**}} & \textbf{69.43±14\textsuperscript{**}} & $44.42±27^{**}$ & $32.60±23^{**}$ & $43.28±31^{**}$ & $32.92±27^{**}$ & $72.20±26$ & $61.77±26$ \\
& TG3K & $37.27±19^{**}$ & $24.57±14^{**}$ & $1.57±5^{**}$ & $0.88±3^{**}$ & \textbf{82.95±6\textsuperscript{**}} & \textbf{71.22±8\textsuperscript{**}} & $67.82±17^{**}$ & $53.50±17^{**}$ & $70.75±20^{**}$ & $57.73±20^{**}$ & $80.32±9$ & $68.04±12$ \\
\midrule
\multirow{2}{*}{Liver} 
& 105US & $23.93±27$ & $16.65±20$ & $39.38±36$ & $31.37±31$ &  $9.24±14$ & $5.52±9$ & $32.69±31$ & $24.20±25$ & $44.31±34$ & $34.85±29$ & \textbf{91.32±0} & \textbf{84.03±0} \\
& AUL & $14.52±15^{**}$ & $8.60±10^{**}$ & $19.78±32^{**}$ & $15.48±26^{**}$ & $28.18±22$ & $18.59±17$ &  $32.12±27$ & $22.69±22^{**}$ &  $32.86±34$ & $25.48±29$ & \textbf{44.02±36} & \textbf{35.79±33} \\
\midrule
\multirow{2}{*}{Prostate} 
& MicroSeg &  $11.62±9^{**}$ & $6.44±6^{**}$ & $23.98±24^{**}$ & $16.25±20^{**}$ & $70.84±18^{**}$ & $57.28±18^{**}$ & $58.67±15^{**}$ & $42.94±14^{**}$ & $53.04±30^{**}$ & $41.10±25^{**}$ & \textbf{88.56±13} & \textbf{81.17±15} \\
& RegPro  &  $16.84±14^{**}$ & $9.89±9^{**}$ & $38.88±32^{**}$ & $29.53±27^{**}$ & $49.22±29^{**}$ & $37.30±25^{**}$ & $42.90±28^{**}$ & $31.26±23^{**}$ & $46.90±31^{**}$ & $36.00±27^{**}$ & \textbf{70.77±31} & \textbf{61.63±29} \\
\midrule
\multirow{1}{*}{Kidney} 
& KidneyUS & $22.22±15^{**}$ & $13.36±10^{**}$ & $48.03±23^{**}$ & $34.55±20^{**}$ & $56.37±18^{**}$ & $41.39±18^{**}$ & $59.86±16^{**}$ & $44.45±15^{**}$ & $64.93±21^{**}$ & $51.10±20^{**}$ & \textbf{73.49±16} & \textbf{60.48±19} \\

\bottomrule
\end{tabular}
\end{table*}

\subsection{Generalization to Unseen Datasets}

To evaluate generalization, we tested all methods on three publicly available ultrasound datasets from unseen domains: breast (BUSBRA), thyroid (TNSCUI), and multifidus muscle (Luminous). UniverSeg has an advantage in that we provided it with a 16-image labelled support set from each unseen dataset before inference.  It should also be noted that, as with the seen dataset, random points within the ground truth mask are provided as spatial prompts for SAMUS.

As shown in Table~\ref{tab:unseen}, our method achieves the best performance across most benchmarks. On the BUSBRA breast dataset, it achieves the top DSC of 79.10\% and IoU of 86.44\%. On the TNSCUI thyroid dataset, our model again outperforms all others, reaching a DSC of $86.44\%$ and IoU of $79.10\%$. Although SAMUS slightly outperforms our method in DSC on the Luminous dataset ($65.36\%$ vs. $64.99\%$), our model achieves the highest IoU of 51.96\%, indicating stronger alignment with the ground truth masks. Notably, both SAMUS and BiomedParse benefit from large-scale training data—30,000 and 3.4 million images, respectively. In contrast, our model is trained on significantly fewer samples and still demonstrates strong generalization across different anatomical structures and imaging conditions, without retraining or task-specific adaptation.

\setlength{\tabcolsep}{1pt}
\begin{table*}[ht]
    \centering
    \scriptsize

    \caption{Quantitative comparison of segmentation performance (DSC and IoU scores) on unseen public ultrasound datasets across different organs using SOTA methods: UniverSeg, BiomedParse, SAMUS, MedCLIP-SAM, MedCLIP-SAMv2, and the proposed method. Results are presented as mean $\pm$ standard deviation. Bolded results indicate the highest score for each comparison.
    Statistical significance was assessed using paired t-tests: $^{**}$  denotes $p < 0.005$ and $^{*}$ denotes $p < 0.01$.}

    \label{tab:unseen}
    \renewcommand{\arraystretch}{1.2}
        \begin{tabular}{llcccccccccccc}
        \toprule
        \multirow{2}{*}{\textbf{Organ}} & \multirow{2}{*}{\textbf{Dataset}} & 
        \multicolumn{2}{c}{\textbf{UniverSeg}} &
        \multicolumn{2}{c}{\textbf{BiomedParse}} &
        \multicolumn{2}{c}{\textbf{SAMUS}} &
        \multicolumn{2}{c}{\textbf{MedCLIP-SAM}} & 
        \multicolumn{2}{c}{\textbf{MedCLIP-SAMv2}} & 
        \multicolumn{2}{c}{\textbf{Ours}} \\
        \cmidrule(lr){3-4} \cmidrule(lr){5-6} \cmidrule(lr){7-8} \cmidrule(lr){9-10} \cmidrule(lr){11-12}\cmidrule(lr){13-14}
        & & \textbf{DSC (\%) $\uparrow$} & \textbf{IoU (\%) $\uparrow$} & \textbf{DSC (\%) $\uparrow$} & \textbf{IoU (\%) $\uparrow$} & \textbf{DSC (\%) $\uparrow$} & \textbf{IoU (\%) $\uparrow$} & \textbf{DSC (\%) $\uparrow$} & \textbf{IoU (\%) $\uparrow$} & \textbf{DSC (\%) $\uparrow$} & \textbf{IoU (\%) $\uparrow$} & \textbf{DSC (\%) $\uparrow$} & \textbf{IoU (\%) $\uparrow$}  \\
        
       \midrule
\multirow{1}{*}{Breast} 
& BUSBRA	& $47.97±27^{**}$ & $35.72±24^{**}$	& $78.79±22^{**}$ & $69.05±23^{**}$	& $75.04±15^{**}$ & $64.80±17^{**}$	& $53.58±25^{**}$ & $40.47±23^{**}$	& $57.82±38^{**}$ & $49.93±35^{**}$	& \textbf{79.10±18} & \textbf{86.44±18} \\
\midrule
\multirow{1}{*}{Thyroid} 
& TNSCUI	& $6.21±7^{**}$ & $3.37±4^{**}$	& $40.54±39^{**}$ & $33.62±34^{**}$	& $63.70±23$ & $57.50±24$	& $63.90±13^{**}$ & $48.17±13^{**}$	& $76.34±12$ & $63.06±14$	& \textbf{86.44±17} & \textbf{79.10±20} \\
\midrule
\multirow{1}{*}{Multifidus Muscle} 
& Luminous	& $25.54±16^{**}$ & $15.69±11^{**}$	& $12.59±18^{**}$ & $7.86±12^{**}$	& \textbf{65.36±13} & $49.99±15$	& $52.27±25^{*}$ & $38.96±22^{**}$	& $62.96±24$ & $49.47±21$	& $64.99±24$ & \textbf{51.96±22} \\
\bottomrule
    \end{tabular}
\end{table*}

\subsection{Results with Different Text Prompts}

Table~\ref{tab:multi_organ_segmentation} presents the segmentation performance (DSC and IoU) for various anatomical structures in the kidney and thyroid, using a range of text prompts. The results demonstrate that prompt phrasing has a considerable impact on segmentation accuracy.
Well-designed prompts that are anatomically accurate and concise, such as \textit{``segment kidney cortex''} or \textit{``segment the thyroid nodules''}, achieved high Dice Similarity Coefficient (DSC) and Intersection over Union (IoU) scores across both organs. For instance, segmenting the kidney capsule with the prompt \textit{``segment kidney capsule''} yielded a DSC of 91.74\% and IoU of 85.10\%, while the thyroid parenchyma using \textit{``segment the glandular portion of the thyroid''} achieved a DSC of 77.07\% and IoU of 64.73\%.
In contrast, prompts that were ambiguous, anatomically irrelevant, or unrelated to the target structure (highlighted in gray in the table) resulted in significantly lower performance. Examples include prompts such as \textit{``highlight the middle phalanges''} or \textit{``segment zones of the uterus''}, which consistently yielded suboptimal scores regardless of the target organ. This drop in performance highlights the model's reliance on the semantic alignment between the prompt and the visual content.
Overall, these results highlight the crucial role of a precise and contextually relevant prompt formulation. To ensure consistency and clinical relevance, we consulted with a radiologist when designing the primary prompt templates used in this study.

\begin{table*}[ht]
\centering
\caption{Segmentation performance (DSC and IoU) for various anatomical structures using different text prompts. Poorly performing prompts are highlighted in gray to illustrate the limitations of ill-conceived or ambiguous phrasing. Results are reported as mean $\pm$ standard deviation.}

\label{tab:multi_organ_segmentation}
\begin{tabular}{>{\raggedright}p{1.5cm} >{\raggedright}p{3.3cm} p{5cm} p{2cm} p{2cm}}
\toprule
\textbf{Organ} & \textbf{Structure} & \textbf{Prompt Text} & \textbf{DSC (\%) $\uparrow$} & \textbf{IoU (\%) $\uparrow$} \\
\midrule

\multirow{28}{*}{\parbox{1.5cm}{Kidney\\(KidneyUS)}}
    & Cortex & segment kidney cortex & 70.01±14 & 55.46±15 \\
    &  & segment renal cortex & 68.17±18 & 54.13±17 \\
    &  & outline kidney cortical tissue & 70.18±13 & 55.52±14 \\
    &  & delineate cortex & 68.25±19 & 54.30±17 \\
    &  & cortical region of the kidney & 17.82±31 & 14.12±25 \\
    &  \cellcolor{gray!20} & \cellcolor{gray!20} highlight the middle phalanges & \cellcolor{gray!20} 44.35±33 & \cellcolor{gray!20} 34.13±26 \\
    &  \cellcolor{gray!20} & \cellcolor{gray!20} segment zones of the uterus & \cellcolor{gray!20} 41.90±37 & \cellcolor{gray!20} 33.78±30 \\

\cmidrule{2-5}
    & Capsule & segment kidney capsule & 91.74±5 & 85.10±8 \\
    &  & segment renal capsule & 91.69±5 & 85.03±7 \\
    &  & outline kidney capsule boundary & 89.72±8 & 82.31±12 \\
    &  & identify capsule & 90.69±8 & 83.77±11 \\
    &  & capsular layer of the kidney & 41.73±43 & 37.57±40 \\
    &  \cellcolor{gray!20} & \cellcolor{gray!20} highlight the middle phalanges & \cellcolor{gray!20} 43.55±33 & \cellcolor{gray!20} 33.80±32 \\
    &  \cellcolor{gray!20} & \cellcolor{gray!20} segment zones of the uterus & \cellcolor{gray!20} 42.86±42 & \cellcolor{gray!20} 37.96±39 \\

\cmidrule{2-5}
    & Medulla & segment kidney medulla & 68.79±18 & 68.79±18 \\
    &  & segment renal medulla & 67.83±20 & 54.18±18 \\
    &  & identify renal medullary area & 41.13±36 & 32.80±30 \\
    &  & outline medulla & 66.18±21 & 52.68±20 \\
    &  & medullary area of kidney & 22.94±33 & 18.12±27 \\
    &  \cellcolor{gray!20} & \cellcolor{gray!20} highlight the middle phalanges & \cellcolor{gray!20} 41.97±35 & \cellcolor{gray!20} 33.18±29 \\
    &  \cellcolor{gray!20} & \cellcolor{gray!20} segment zones of the uterus & \cellcolor{gray!20} 40.24±37 & \cellcolor{gray!20} 32.52±31 \\

\cmidrule{2-5}
    & Central Echo Complex & segment central echo complex of kidney & 67.98±15 & 53.37±16 \\
    &  & segment renal central echo complex & 68.45±14 & 53.77±16 \\
    &  & delineate central echoes of kidney & 64.25±21 & 50.46±20 \\
    &  & outline central echoes & 63.59±21 & 49.80±20 \\
    &  & renal central echoes & 65.71±18 & 51.39±18 \\
    &  \cellcolor{gray!20} & \cellcolor{gray!20} highlight the middle phalanges & \cellcolor{gray!20} 49.14±31 & \cellcolor{gray!20} 37.89±26 \\
    &  \cellcolor{gray!20} & \cellcolor{gray!20} segment zones of the uterus & \cellcolor{gray!20} 42.14±35 & \cellcolor{gray!20} 33.05±28 \\
\midrule
\multirow{14}{*}{\parbox{1.5cm}{Thyroid\\(TN3K,\\TG3K)}}
    & Parenchyma & segment the thyroid parenchyma & 70.32±26 & 59.22±24 \\
    &  & segment the glandular portion of the thyroid & 77.07±15 & 64.73±16 \\
    &  & outline the thyroid tissue & 74.29±20 & 62.37±20 \\
    &  & delineate parenchyma & 31.64±38 & 26.36±33 \\
    &  & glandular area of the thyroid & 75.32±19 & 63.39±19 \\
    &  \cellcolor{gray!20} & \cellcolor{gray!20} highlight the middle phalanges & \cellcolor{gray!20} 55.11±37 & \cellcolor{gray!20} 46.34±32 \\
    &  \cellcolor{gray!20} & \cellcolor{gray!20} segment zones of the uterus & \cellcolor{gray!20} 50.88±25 & \cellcolor{gray!20} 49.46±23 \\
\cmidrule{2-5}
    & Nodule & segment the thyroid nodules & 75.90±21 & 65.02±22 \\
    &  & segment the focal lesions in the thyroid & 75.10±21 & 64.06±22 \\
    &  & identify the thyroid lump & 72.13±26 & 61.57±25 \\
    &  & outline nodules & 72.04±27 & 61.83±26 \\
    &  & hypoechoic region in the thyroid & 33.59±41 & 29.54±37 \\
    &  \cellcolor{gray!20} & \cellcolor{gray!20} highlight the middle phalanges & \cellcolor{gray!20} 51.96±30 & \cellcolor{gray!20} 45.21±28 \\
    &  \cellcolor{gray!20} & \cellcolor{gray!20} segment zones of the uterus & \cellcolor{gray!20} 45.94±33 & \cellcolor{gray!20} 43.78±30 \\
\bottomrule
\end{tabular}
\end{table*}

\subsection{Qualitative Results}
\begin{figure*}[ht]
    \centering
    \includegraphics[width=\textwidth]{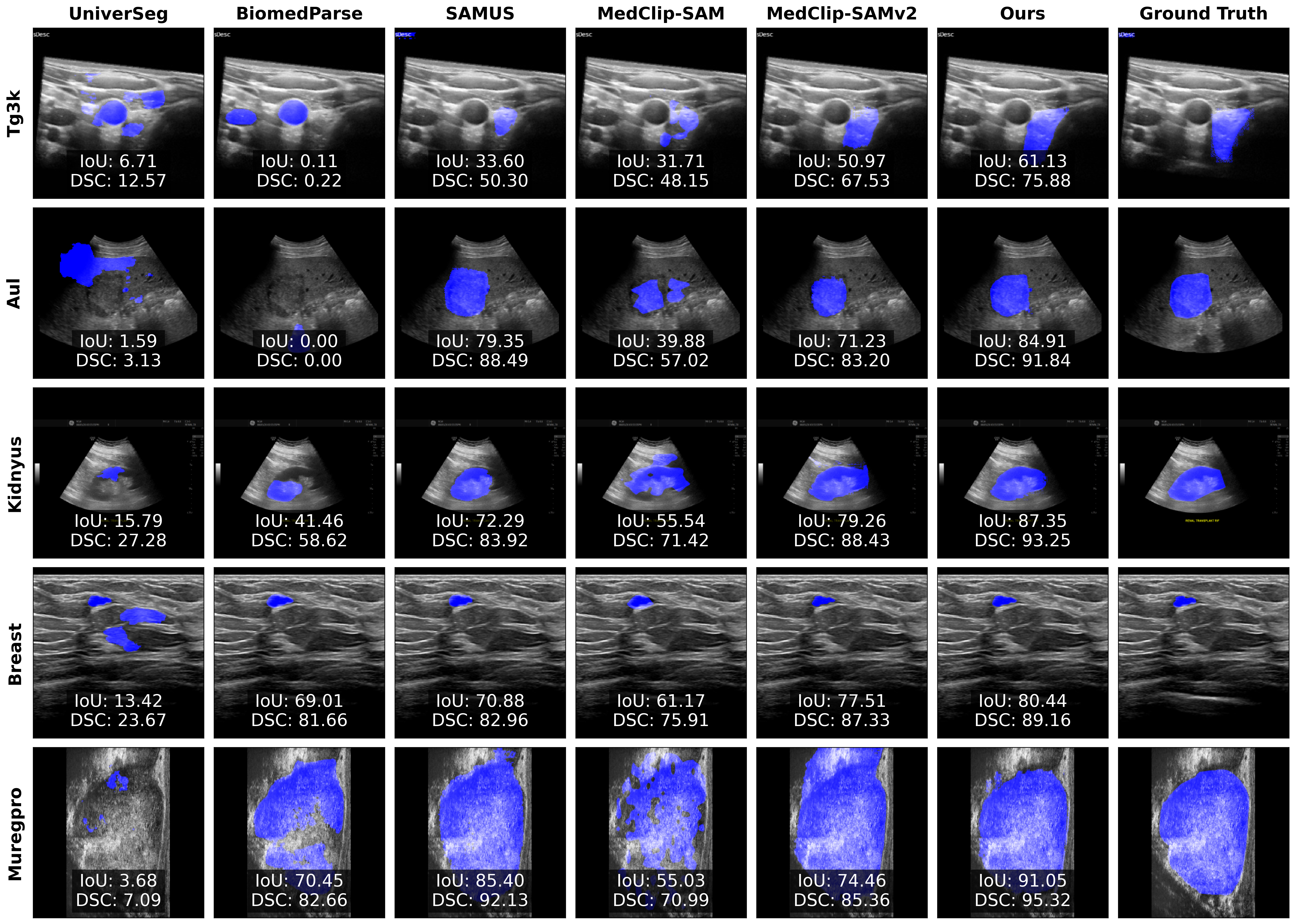}
    
    \caption{Qualitative comparison of segmentation results on various seen ultrasound datasets. Each row represents a distinct dataset: Tg3k (Thyroid), Aul (Liver), Kidneyus (Kidney), BrEast (Breast), and Muregpro (Prostate). Columns show predictions from UniverSeg, BiomedParse, SAMUS, MedCLIP-SAM, MedCLIP-SAMv2, our method, and the Ground Truth masks. The predicted segmentation masks are overlaid in blue. Corresponding DSC and IoU scores are provided for each prediction to illustrate segmentation performance quantitatively.}
    \label{fig:seen}
\end{figure*}

\begin{figure*}[ht]
    \centering
    \includegraphics[width=\textwidth]{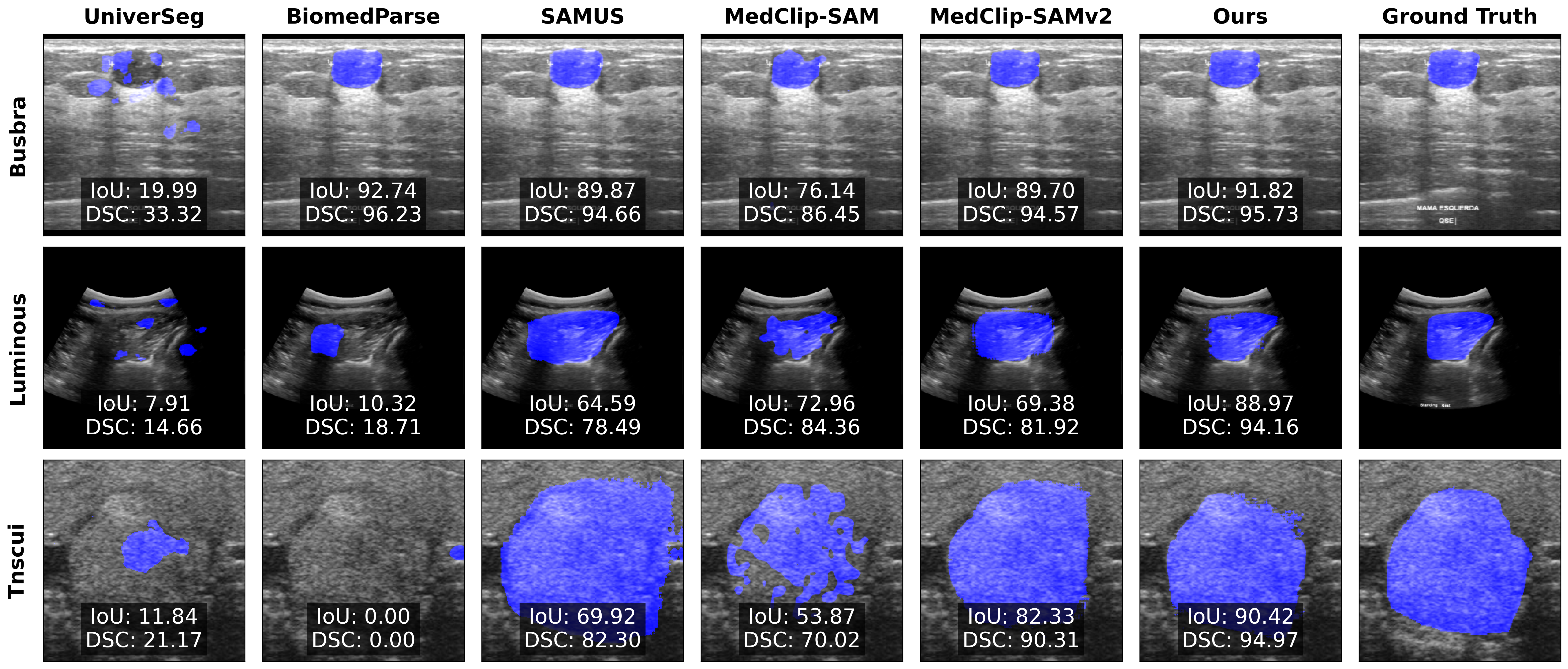} 
    
    \caption{Qualitative comparison of segmentation results on various unseen ultrasound datasets. Each row represents a distinct dataset: Busbra (Breast), Luminous (Lower Back Muscle), and Tnscui (Thyroid). Columns show predictions from UniverSeg, BiomedParse, SAMUS, MedCLIP-SAM, MedCLIP-SAMv2, our method, and the Ground Truth masks. The predicted segmentation masks are overlaid in blue. Corresponding DSC and IoU scores are provided for each prediction to illustrate segmentation performance quantitatively.}

    \label{fig:unseen}
\end{figure*}

Figures~\ref{fig:seen} and~\ref{fig:unseen} present qualitative comparisons of segmentation results on both seen and unseen ultrasound datasets. Each row represents a different dataset covering various organs. Columns show predictions from competing SOTA methods, UniverSeg, BiomedParse, SAMUS, MedCLIP-SAM, MedCLIP-SAMv2, alongside the proposed method (Ours) and the corresponding ground truth masks.
For seen datasets (Fig.~\ref{fig:seen}), the proposed method delivers the most accurate boundary delineation with fewer false positives, especially for complex structures such as the prostate and liver. This superior performance is evident in the higher DSC and IoU scores compared to other methods.

Similarly, for unseen datasets (Fig.~\ref{fig:unseen}), our method consistently achieves better segmentation quality and generalizes well to previously unseen data. Notably, in challenging cases like the Luminous (Lower Back Muscle) and TNSCUI (Thyroid) datasets, the proposed model preserves anatomical structure and produces minimal over-segmentation or under-segmentation, reflecting strong domain generalization capabilities without task-specific fine-tuning.

\subsection{Runtimes}
Inference runtimes were evaluated on an NVIDIA TITAN V GPU (12\,GB). As summarized in Table~\ref{tab:runtime}, the proposed method requires an average of \textbf{0.33\,s} per image, outperforming BiomedParse (0.49\,s), SAMUS (0.67\,s), MedCLIP-SAM (3.05\,s), and MedCLIP-SAMv2 (2.73\,s). UniverSeg achieved the fastest runtime (0.21\,s) due to its lightweight, task-specific design without text prompt conditioning; however, it underperforms and further requires a support set of size 16 for all tasks (even the unseen dataset). Our method offers superior flexibility and segmentation accuracy at a modest computational cost.

\begin{table}[ht]
    \centering
    \caption{Average inference time per image (mean ± standard deviation) over 10 runs on NVIDIA TITAN V with 12\,GB of RAM.
    }
    \label{tab:runtime}
    \begin{tabular}{lc}
        \toprule
        \textbf{Method} & \textbf{Runtime (s) $\downarrow$} \\
        \midrule
        {UniverSeg} & {0.21±0.002} \\
        BiomedParse & 0.49±0.045 \\
        SAMUS & 0.67±0.005 \\
        MedCLIP-SAM & 3.05±0.099 \\
        MedCLIP-SAMv2 & 2.73±0.085 \\
        {Ours} & {0.33±0.032} \\
        \bottomrule
    \end{tabular}
\end{table}

\section{Discussion}
The results of Table~\ref{tab:ablation} show that although MedSAM~\cite{ma2024segment} is explicitly fine-tuned on medical images, its core architecture is still the first-generation SAM~\cite{kirillov2023segment}, which uses a single-scale ViT encoder and therefore has to trade off between global context and the high-frequency speckle patterns. SAM2 replaces that backbone with a dual-scale ViT that processes high-resolution local tokens in parallel with coarse global tokens. This design captures texture (such as edges and speckles) and long-range anatomical context simultaneously, qualities that are important for ultrasound segmentation but are only partially learned in MedSAM. Moreover, SAM2 uses a substantially larger pre-training dataset and a refined mask decoder~\cite{ravi2024sam}.

BiomedParse and SAMUS benefited from significantly larger pre-training datasets, yet our model, fine-tuned on a much smaller dataset, consistently achieves the best segmentation results. It should also be noted that SAMUS requires a point prompt at test time, and in our evaluation, we supplied a randomly chosen point inside each ground-truth mask. Our method, in contrast, operates using solely text and receives no spatial guidance, making the evaluation deliberately conservative and underlining the method’s robustness. These findings show that Grounding DINO encoder pre-trained almost entirely on natural-image/text pairs can be tuned to the ultrasound domain with minimal data. In other words, LoRA fine-tuning is enough to align visual and language cues. 
When comparing the poor results of Grounding DINO + SAM2 to the excellent results of Fine-tuned Grounding DINO + SAM2 in Table~\ref{tab:ablation}, it is evident that the SAM2 decoder is capable of high-quality ultrasound segmentation when an accurate bounding box is available. In other words, this implies that fine-tuning is required solely for Grounding DINO, with no modifications needed for SAM2.  While our preliminary results show that it is not necessary to fine-tune SAM2 to the ultrasound domain, we will explore this fine-tuning in future work.

To evaluate generalization, we tested the model on three datasets that were entirely unseen during fine-tuning. In particular, the LoRA fine-tuning stage contained no musculoskeletal data, yet the model still produced accurate masks on the Luminous multifidus-muscle dataset, showing its robustness to domain shift. This contrasts sharply with the recent multi-organ foundation model proposed by Chen \textit{et al.}~\cite{chen2025multi}, which must be trained jointly on several organs to acquire organ-invariant features. By accepting free-form text prompts, our system can segment previously unseen anatomies or pathologies without any additional retraining, demonstrating the practical value of an open-vocabulary, prompt-driven approach for future clinical applications.

Adapting large foundation models to the ultrasound domain is computationally and memory-intensive, but our method addresses this with LoRA fine-tuning for efficiency. Rather than updating all model weights, we fine-tune only a small fraction of parameters in the Grounding DINO pipeline by inserting lightweight LoRA modules. To the best of our knowledge, this is the first time a VLM has been successfully fine-tuned to the ultrasound domain using LoRA. 

A strength of our method is its robustness to differences in text prompts. The model uses Grounding DINO’s open-vocabulary, which is built on a VLM, to interpret arbitrary text inputs and localize the corresponding region in ultrasound images. We observed that synonyms and varying descriptions provide similar segmentation results, indicating a high tolerance for language variations. This flexibility is important in clinical practice, as different practitioners may describe the same anatomy in different terms. 

Our approach has several advantages over recent SAM-based and foundation models in medical imaging. Compared to MedSAM ~\cite{ma2024segment}, our model avoids the costly full-model retraining and the dependence on manual point or box spatial prompts. Compared to SAMUS~\cite{lin2024beyond}, which introduced a parallel CNN branch and an automatic prompt generator to adapt SAM for ultrasound segmentation, our approach remains architecturally simpler by simply pairing a text-conditioned detector (i.e., DINO) with SAM2. SAMUS’s learned prompts are task-specific (trained for a fixed set of ultrasound targets), but our approach can handle any structure on the fly, making it more extensible. Finally, relative to Chen~\textit{et al.}~\cite{chen2025multi}, our model does not rely on predefined organ classes or hard-coded priors. Our model can flexibly segment structures with only a descriptive text prompt, a significant advantage in scenarios where one needs to segment an arbitrary anatomy that a model like Chen~\textit{et al.}~\cite{chen2025multi} was never explicitly trained on. 

Our method’s understanding of a text prompt is limited by the alignment between visual features and text learned during training. Because the text encoder (from Grounding DINO) was initially trained on natural images and captions, some medical terms or subtle features of ultrasound images may not be represented in the model. Although we mitigated this by fine-tuning Grounding DINO on ultrasound data, the exact boundary of an organ might be harder to segment for SAM2 if the model hasn’t seen similar images. These challenges suggest that while language grounding is a powerful tool, curation of a rich ultrasound dataset is needed to ensure the model understands medical terminologies and cues in ultrasound images.
Future work will explore such data curation and will incorporate prompt tuning~\cite{koleilat2025biomedcoop} to further enhance the performance.

\section{Conclusion}
We proposed a Grounding DINO-based VLM integrated with SAM2 for ultrasound segmentation across multiple organs and imaging scenarios. Evaluated on 18 public ultrasound datasets, our method consistently outperforms recent approaches, including MedCLIP-SAM, MedCLIP-SAMv2, SAMUS, and UniverSeg on both seen and unseen data. Our model is fully automated and doesn't require any spatial prompts. It also shows strong generalization performance on three unseen datasets. Finally, it achieves a competitive average runtime of 0.33 seconds per image on an affordable GPU, making it suitable for real-time clinical use.
\section{Acknowledgment}
Funded by Natural Sciences and Engineering Research Council of Canada (NSERC) Discovery and Government of Canada’s New Frontiers in Research Fund (NFRF) [NFRFE-2022-00295]  grants. We would like to thank Dr. Khashayar Rafatzand for invaluable discussions and NVIDIA for the donation of the GPU. 

\bibliographystyle{IEEEtran}
\bibliography{ref.bib}

\end{document}


\maketitle

\section{DICOM images}
\begin{figure*}[ht]
    \centering
    \includegraphics[width=.9\linewidth,height=.5\linewidth]{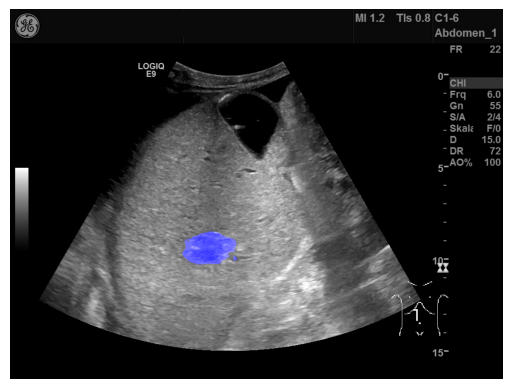}
    \caption{Example input image with text prompt of ``Segment the gallbladder lumen, liver parenchyma, and focal hepatic lesion'' from the test set. The image contains both relevant anatomical content and irrelevant border regions that include patient information and other text commonly found in  DICOM images.}
    \label{fig:dicom_image_seen}
\end{figure*}

\begin{figure*}[ht]
    \centering
    \includegraphics[width=.9\linewidth,height=.5\linewidth]{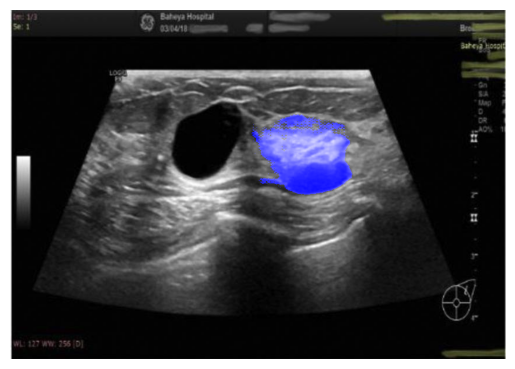}
    \caption{Example input image with text prompt of ``Segment all breast lesions'' from the unseen dataset. The image contains both relevant anatomical content and irrelevant border regions that include patient information and other text commonly found in  DICOM images.}
    \label{fig:dicom_image_unseen}
\end{figure*}

The examples in Figures~\ref{fig:dicom_image_seen} and~\ref{fig:dicom_image_unseen} demonstrate that the model is capable of accurate segmentation from full-field-of-view DICOM-derived ultrasound images, without requiring spatial prompts or manual region selection. Although the input images were normalized to a resolution of 800x800 pixels, this step was performed for consistency and does not isolate target structures. The model successfully segments relevant anatomical regions despite the presence of non-target tissues and imaging artifacts, indicating robustness to irrelevant content in the input image. Notably, the example in Figure~\ref{fig:dicom_image_unseen}, which comes from an unseen breast ultrasound dataset, shows that the model generalizes well to new domains and continues to perform accurate segmentation based solely on the text prompt.

\section{Multi-Organ segmenation}
In cases where an image contains multiple instances of the target—such as several benign nodules—the model successfully detects and segments each instance separately. As shown in Figure~\ref{fig:multi_organ}, our method outputs multiple bounding boxes and corresponding masks in response to a single text prompt. This demonstrates the model’s ability to localize and segment multiple objects of the same category within a single image, highlighting its potential for comprehensive multi-target ultrasound analysis.

\begin{figure}[H]
    \centering
    \includegraphics[width=.9\linewidth]{TUFFC-Review-2025-07-17/images/Multi-target/busi_benign(83).png}
    \caption{Example of multi-organ detection and segmentation on a breast ultrasound image. The model identifies and segments multiple benigns within the same image using a single prompt. The blue masks indicate successful segmentation of multiple targets.}
    \label{fig:multi_organ}
\end{figure}